\documentclass[10pt,twocolumn,letterpaper]{article}

\usepackage[pagenumbers]{cvpr}

\definecolor{cvprblue}{rgb}{0.21,0.49,0.74}
\usepackage[pagebackref,breaklinks,colorlinks,allcolors=cvprblue]{hyperref}
\usepackage{multirow}
\usepackage{algorithm}
\usepackage{algpseudocode}
\usepackage{pifont}
\usepackage[dvipsnames,table]{xcolor}
\definecolor{darkgreen}{RGB}{0,150,0}

\newcommand{\ModuleName}{\texttt{IP-CIR}}

\title{Imagine and Seek: Improving Composed Image Retrieval with an Imagined Proxy}

\author{
    You Li \qquad
    Fan Ma \quad
    Yi Yang$^\dag$ \\
    ReLER, CCAI, Zhejiang University, Zhejiang, China\\
    {\small $^\dag$ Corresponding author} \\
    {\tt\small {\{uli2000, mafan, yangyics\}}@zju.edu.cn} \\
}

\begin{document}
\maketitle
\begin{abstract}

The Zero-shot Composed Image Retrieval (ZSCIR) requires retrieving images that match the query image and the relative captions.
Current methods focus on projecting the query image into the text feature space, subsequently combining them with features of query texts for retrieval. However, retrieving images only with the text features cannot guarantee detailed alignment due to the natural gap between images and text.
In this paper, we introduce \textbf{I}magined \textbf{P}roxy for CIR (\ModuleName), a training-free method that creates a proxy image aligned with the query image and text description, enhancing query representation in the retrieval process.
We first leverage the large language model’s generalization capability to generate an image layout, and then apply both the query text and image for conditional generation.
The robust query features are enhanced by merging the proxy image, query image, and text semantic perturbation. Our newly proposed balancing metric integrates text-based and proxy retrieval similarities, allowing for more accurate retrieval of the target image while incorporating image-side information into the process.
Experiments on three public datasets demonstrate that our method significantly improves retrieval performances. We achieve state-of-the-art (SOTA) results on the CIRR dataset with a Recall@K of 70.07 at K=10. Additionally, we achieved an improvement in Recall@10 on the FashionIQ dataset, rising from 45.11 to 45.74, and improved the baseline performance in CIRCO with a mAPK@10 score, increasing from 32.24 to 34.26. 
\end{abstract}    
\section{Introduction}
\label{sec:intro}

\textit{Imagination is the wellspring of human creativity.} When you see a picture of an adorable cat, you might envision it playing with other cats, wearing cool sunglasses and a cape, or floating in a space capsule. This ability to imagine and pursue beautiful visions drives continuous advancement in fields like image generation~\cite{stablediffusion,ddim,ddpm,eDiff-I,glide,DALL-E2,pydiff,3dis,caphuman} and image retrieval~\cite{cirr,circo,fashioniq,pic2word,searle,isearle,ldre,lincir,magiclens,anys}.

\begin{figure}[tb]
    \setlength{\abovecaptionskip}{-0.cm}
    \setlength{\belowcaptionskip}{-0.cm}
    \begin{center}
        \includegraphics[width=0.48\textwidth]{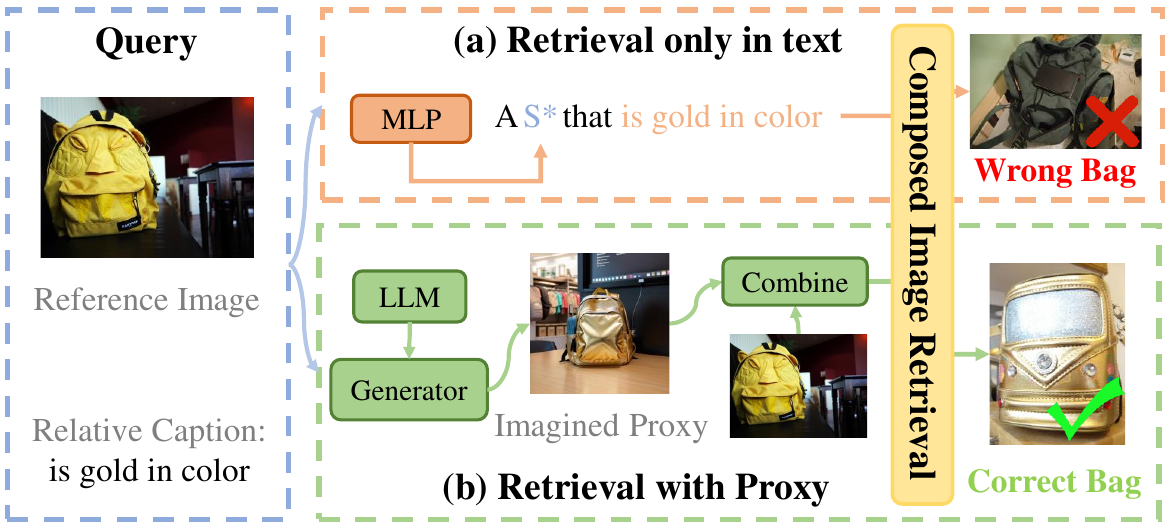}
        
    \end{center}

    \caption{ \textbf{Comparison of composed image retrieval between text-only retrieval and our methods.} Traditional methods perform retrieval only in the text space, where CLIP text features may overlook some important semantic information. In contrast, our approach generates imagined proxy features, providing additional information that is often overlooked in text-only retrieval, thereby improving retrieval accuracy. } 
    
    \label{fig:fig_1}
    \vspace{-8pt}
  \end{figure}

\textit{This rich imagination gains new significance in Composed Image Retrieval} (CIR) \cite{cirr}, where users express their imagination of the query image through text, searching for content that aligns with their preferences. Specifically, the CIR task requires users to retrieve target images from a database that match both the content of a query image and relative caption. With text modality to flexibly control retrieval targets, \textit{ CIR enables the search process akin to the user’s specific imagination after viewing the query image}. This \textbf{multi-modal, imagination-like} retrieval task greatly expands the applications of image retrieval, with wide-ranging uses in fields such as visual search~\cite{visualsearch1}, object localization~\cite{local1,local2}, and re-identification~\cite{reid1,reid2}.

Traditional solutions for CIR~\cite{cir1,cir2,cir3,cir4,cir5,cir6,cir7,cir8} are constrained by the scarcity of training data (e.g, 46.6k for FashionIQ~\cite{fashioniq} dataset and 28.8k for CIRR~\cite{cirr} dataset), as collecting suitable triplets (query image  $x_q$, relevant text $x_c$, and target image $x_t$) from the web incurs high labor costs. Therefore, we focus on the ZS-CIR task, aiming to bypass the need for training on triplet datasets and achieve generalized CIR retrieval capabilities. 
As shown in Fig.\ref{fig:fig_1} (a), traditional ZS-CIR methods typically employ lightweight projection models to map CLIP’s~\cite{CLIP} image features into the text features~\cite{searle} for retrieval. \textbf{\textit{The transformation inevitably loses certain image information, and the coarse, easily confusable text features fall short in fine-grained, complex imagination scenes.}} Although recent approaches attempt to leverage large language models (LLMs)~\cite{gpt4,qwen} to generate various descriptions of target images~\cite{vbl,ldre,magiclens}, they overlook the potential for direct imagination on the image side. As shown in Fig.\ref{fig:fig_1} (a), the text features did not effectively capture the attribute `golden', resulting in suboptimal retrieval results.

The rapid advancement of controllable generative models~\cite{gligen,instancediffusion,migc++,aae,anydoor,boxdiff,imagen,reco} has empowered users with the freedom to create images that align with their imagined visions. Generating images that match both the content of query images and relative captions aligns closely with the goals of composed retrieval tasks, and the imagined images (we called \textbf{Retrieval Proxy}) could provide additional information—such as style, instance attributes, and spatial relationships—that is often overlooked in text-based retrieval with complex captions. This raises an intriguing question: \textit{\textbf{How could we harness the imaginative power of generative models to improve retrieval performance?}} 

We thus propose \ModuleName\ (Imagined Proxy for CIR), a training-free, plug-and-play method that harnesses the power of imagination for any retrieval method. Firstly, by utilizing LLMs to understand both the image content and the relative captions, we generate a suitable image layout. Using controllable generation methods, we create high-quality, fine-grained proxy images. Next, we compose the proxy images, query images as well as semantic perturbation into robust proxy features that are more suitable for retrieval. Finally, we propose a balance metric to combine the proxy and text retrieval similarities, allowing the integration of our method with any retrieval methods.

We conduct experiments on three public datasets: the CIRR~\cite{cirr}, CIRCO~\cite{circo}, and FashionIQ~\cite{fashioniq}. We combine our method with several baselines, the observed performance improvements confirm that our method supplements the retrieval process with valuable information and could enhance the performance of any other retrieval methods.

Our contributions are summarized as follows:

\begin{itemize}
  \item [1)] 
  We propose using generative models' imagination capabilities to create proxy images that align with the content of the query image and relative captions to improve retrieval performance. 
  \item [2)]
  We introduce a proxy image generation framework. Using the LLM-generated object layout and the query image, we create imagined proxies through MIGC++.
  \item [3)]
  We propose integrating proxy images with the query image and semantic perturbation into more robust proxy features and introducing a balance metric to combine the proxy and baseline similarities.
  \item [4)]
  We conducted experiments on three public CIR datasets. The improvement over baseline results, as well as achieving SOTA results demonstrates the potential of constructing imagined proxies in retrieval.
  
\end{itemize}
\section{Related work}
\label{sec:related_work}

\begin{figure*}[tb]
    \setlength{\abovecaptionskip}{-0.cm}
    \setlength{\belowcaptionskip}{-0.cm}
    \begin{center}
        \includegraphics[width=1.0\textwidth]{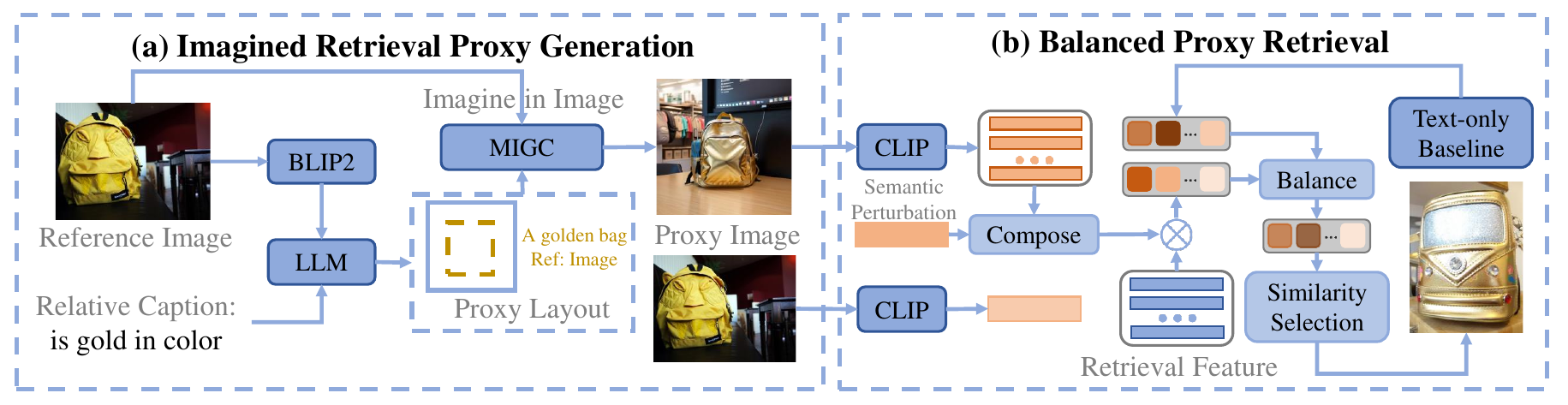}
        
    \end{center}

    \caption{ \textbf{Overview of our method.} (a) represents our imagined proxy generation process. We use LLM to analyze the BLIP2-generated query image captions and the relative captions and infer the proxy layout. We then use the controllable generator to imagine the proxy images. (b) represents our process of constructing a robust proxy feature, and balancing the text and proxy similarities. We integrate proxy features, query image features as well as semantic perturbations into a robust proxy feature, and propose a balance metric for retrieval.} 
    
    \label{fig:fig_2}
  \end{figure*}

\subsection{Zero-shot Composed Image Retrieval}

Early work on ZS-CIR, such as Pic2word~\cite{pic2word}, SEALRE~\cite{searle}, attempted to project the original image into the text space using lightweight projection models. This was often done within fixed templates (e.g., ``a photo of [\$] with cond"), where the query images and relative captions were combined. LinCIR~\cite{lincir} further advanced this approach by defining keywords as adjectives and nouns in the text, thus overcoming the limitations of fixed templates and training more powerful projection models. In addition, some methods have embraced LLMs~\cite{gpt4,qwen}. For instance, CIReVL~\cite{vbl} proposed generating captions for reference images and then using LLMs to reassemble them into target texts modified by relative captions for retrieval. LDRE~\cite{ldre} took this further by using LLMs to construct various edited captions and combining them based on similarity, addressing the issue of fuzzy retrieval. While these methods improved retrieval accuracy by introducing images into the text space, they all overlooked the potential of imagining proxy images to provide additional information in image size.

\subsection{Controllable Text-to-Image Generation}

In recent years, with the development of diffusion models~\cite{ddpm,ddim}, high-quality images that meet specific requirements can now be generated. Using mainstream generators such as Stable Diffusion~\cite{stablediffusion}, users can generate images aligned with text descriptions. Although these foundational generators offer a certain level of control, their performance in fine-grained attribute and orientation control is limited, primarily due to the attribute ambiguity in the CLIP features of the text~\cite{migc}. As a result, some efforts have focused on leveraging layout information to enhance control. For example, GLIGEN~\cite{gligen} introduces Gated-Self-Attention to combine positional feature information, thereby improving control over positioning. InstanceDiffusion~\cite{instancediffusion} further improves attribute control on this foundation. Meanwhile, MIGC~\cite{migc} adopts a divide-and-conquer approach, breaking down complex instance generation into multiple single-instance generations and incorporating Enhancement Attention and Layout Attention to improve both positional and attribute control. MIGC++~\cite{migc++} further integrates ELITE~\cite{elite} to transform images into CLIP-based text features and introduces an additional EA layer to process information from the reference image modality, thereby incorporating image modality control in the generation process. In this paper, we aim to utilize the fine-grained control capabilities of MIGC to display certain attributes and other information that are difficult to clearly express in text at a fine-grained level within the image itself. This provides supplementary information during the retrieval process.

\section{Method}
\label{sec:method}

In this section, we will introduce our \ModuleName, which attempts to imagine an appropriate proxy image within the image feature space, balance, and combine the retrieval results from the image side with those from the text side. Specifically, in section \ref{sec:overview}, we introduce the overview of our method, in section \ref{sec:proxy}, we will describe how to construct an imagined retrieval proxy for each query. In section \ref{sec:ensemble}, 
we will further explain how to organize and integrate the information from the imagined retrieval proxy and present how to combine the retrieval information from both the image and text feature spaces.

\begin{table*}[!htb]
\setlength\tabcolsep{4pt}
\setlength{\abovecaptionskip}{0.cm}
    \setlength{\belowcaptionskip}{0.cm}
  \centering

  \begin{tabular}{c c | c  c c c | c c c c| c c c}
    \toprule
    \multicolumn{2}{c|}{\textbf{Benchmark}} & \multicolumn{4}{c|}{\textbf{CIRCO}} & \multicolumn{7}{c}{\textbf{CIRR}} \\

    \midrule
    
    \multicolumn{2}{c|}{\textbf{Metric}} & \multicolumn{4}{c|}{\textbf{mAP@K}} & \multicolumn{4}{c|}{\textbf{Recall@K}} & \multicolumn{3}{c}{\textbf{Recall$_{Subset}$@K}}  \\
    \midrule
    
     Backbone &  Method &  k=5 & k=10 &  k=25 & k=50 & k=1 & k=5 &  k=10 & k=50 & k=1 & k=2 & k=3  \\
    \midrule
    \multirow{6}{*}{ViT-L/14} & SEARLE & 11.68 & 12.73 & 14.33 & 15.12 & 24.24 & 52.48 & 66.29 & 88.84 & 53.76 & 75.01 & 88.19 \\
     & SEARLE-OTI & 10.18 & 11.03 & 12.72 & 13.67 & 24.87 & 52.31 & 66.29 & 88.58 & 53.80 & 74.31 & 86.94  \\
     & CIReVL & 18.57 & 19.01 & 20.89 & 21.80 & 24.55 & 52.31 & 64.92 & 86.34 & 59.54 & 79.88 & 89.69  \\
     &  LDRE & 23.35 & 24.03 & 26.44 & 27.50 & 26.53 & 55.57 & 67.54 & 88.50 & 60.43 & 80.31 & 89.90  \\
     &  \textbf{LDRE + \ModuleName} & \textbf{26.43} & \textbf{27.41} & \textbf{29.87} & \textbf{31.07} & \textbf{29.76} & \textbf{58.82} & \textbf{71.21} & \textbf{90.41} & \textbf{62.48} & \textbf{81.64} & \textbf{90.89}  \\
     &  (vs. LDRE) & \textcolor{darkgreen}{+3.08} & \textcolor{darkgreen}{+3.38} & \textcolor{darkgreen}{+3.43} & \textcolor{darkgreen}{+3.57} & \textcolor{darkgreen}{+3.23} & \textcolor{darkgreen}{+3.25} & \textcolor{darkgreen}{+3.67} & \textcolor{darkgreen}{+1.91} & \textcolor{darkgreen}{+2.05} & \textcolor{darkgreen}{+1.33}  & \textcolor{darkgreen}{+0.99}  \\
    \midrule
    \multirow{7}{*}{ViT-G/14} &  CIReVL & 26.77 & 27.59 & 29.96 & 31.03 & 34.65 & 64.29 & 75.06 & 91.66 & 67.95 & 84.87 & 93.21 \\
    
    & LinCIR & 20.34 & 21.85 & 23.98 & 25.14 & 34.75 & 64.12 & 75.93 & 93.42 & 62.36 & 81.78 & 91.28 \\
    
    & LinCIR + \ModuleName & 25.70 & 26.64 & 29.09 & 30.13 & 35.37 & 64.70 & 76.15 & 93.71 & 62.58 & 81.74 & 91.35 \\
    & (vs. LinCIR) & \textcolor{darkgreen}{+5.40} & \textcolor{darkgreen}{+4.79} & \textcolor{darkgreen}{+5.11} & \textcolor{darkgreen}{+4.99} & \textcolor{darkgreen}{+0.62} & \textcolor{darkgreen}{+0.58} & \textcolor{darkgreen}{+0.22} & \textcolor{darkgreen}{+0.29} & \textcolor{darkgreen}{+0.22} & \textcolor{red}{-0.04} & \textcolor{darkgreen}{+0.07} \\
    & LDRE & 31.12 & 32.24 & 34.95 & 36.03 & 36.15 & 66.39 & 77.25 & 93.95 & 68.82 & 85.66 & 93.76 \\
    & \textbf{LDRE + \ModuleName} & \textbf{32.75} & \textbf{34.26} & \textbf{36.86} & \textbf{38.03} & \textbf{39.25} & \textbf{70.07} & \textbf{80.00} & \textbf{94.89} & \textbf{69.95} & \textbf{86.87} &  \textbf{94.22} \\
    & (vs. LDRE)& \textcolor{darkgreen}{+1.63} & \textcolor{darkgreen}{+2.02} & \textcolor{darkgreen}{+1.91} & \textcolor{darkgreen}{+2.00} & \textcolor{darkgreen}{+3.10} & \textcolor{darkgreen}{+3.68} & \textcolor{darkgreen}{+2.75} & \textcolor{darkgreen}{+0.94} & \textcolor{darkgreen}{+1.13} & \textcolor{darkgreen}{+1.21} & \textcolor{darkgreen}{+0.46} \\

    \bottomrule
  \end{tabular}
  \caption{ Quantitative results in CIRCO and CIRR datasets.}
  \label{tab:ZSCIR}
  \vspace{5pt}
\end{table*}

\subsection{Overview}
\label{sec:overview}

The overall structure of our framework is shown in Fig.\ref{fig:fig_2}. First, we generate the imagined proxy image for retrieval. The proxy image needs to align with the content of the query image and be as consistent as possible with the description in the relative captions. Therefore, we \textbf{first generate a reasonable object layout} and then use a controllable generation model to \textbf{imagine this layout as a concrete image}. Specifically, we use BLIP2\cite{blip2} to generate a set of captions for each query. Combine with the relative captions, we instruct the LLM to reason out a suitable spatial layout of the target image, so that, with \textbf{as much accuracy as possible} in terms of \textit{\textbf{scene, instance attributes, and quantity}} with relative caption, we can imagine a proxy image of the retrieval target to provide additional information for the retrieval process. In Fig.\ref{fig:fig_2} (b), we compose the imagined proxy with query image and text features, forming a robust retrieval proxy. The similarity of robust proxy is further balanced with other baselines' similarity, improving the retrieval accuracy.

\subsection{Imagined Retrieval Proxy Generation}
\label{sec:proxy}

Text retrieval features may struggle to precisely capture important semantics in the relevant text and the query image, especially when the caption is complex. \textbf{Generating an image that meets the retrieval requirements}, which we refer to as \textbf{imagining the retrieval proxy}, could provide more valuable details to the retrieval process.

\noindent \textbf{Imagine Proxy Layout.} The first step of imagining a suitable proxy relies on \textit{envisioning the overall layout}, which requires a clear understanding of the semantic content and relationship between the query images and relative captions. We thus leverage the robust interpretive and reasoning abilities of LLMs\cite{gpt4,qwen} to generate a target proxy layout. As illustrated in Fig.\ref{fig:fig_2} (a), we organize the BLIP2 generated query image captions with the relative caption into the fixed form \textit{`Given an image of \{caption\}, we show \{rule\}'}, based on which the LLM is instructed to infer the reasonable overall scene and the layout of the proxy image. The layout includes a detailed description of each instance as well as the bounding box (bbox) coordinates. Besides, certain instances may require detailed visual information from text or image modality. For instance, when retrieving an image of a dog wearing a hat, the attributes of the dog in the proxy image come from the dog in the query image, while the attributes of the hat come from the text modality information. We thus require the LLM to determine the reference modality for each instance.

\noindent \textbf{Proxy Image Generation.} The next step is to \textit{imagine the proxy} based on the generated layout. Since the CIR task is a \textbf{fuzzy retrieval task}~\cite{ldre}, this \textit{reduces the need to maintain the query image's instance ID during the generation process}. So we follow the idea of MIGC++~\cite{migc++}, incorporating the query image features transformed by ELITE~\cite{elite} into MIGC, thereby introducing some degree of the query image information into the Proxy image. Besides, an instance can be influenced by both the reference image and the relative caption. For example, if a user envisions a hat in the query image with a specific pattern, the hat should retain its original shape but include the target pattern. So, we copy the layout of the instance with reference image modality and render it based on both image and text, ensuring the proxy image reflects the text’s modifications.

\begin{table*}[!htb]
\setlength\tabcolsep{4pt}
\setlength{\abovecaptionskip}{0.cm}
    \setlength{\belowcaptionskip}{0.cm}
  \centering

  \begin{tabular}{c c  c c c c  c c c c}
    \toprule
    \multicolumn{2}{c}{\textbf{Type}} & \multicolumn{2}{c}{\textbf{Shirt}} & \multicolumn{2}{c}{\textbf{Dress}} & \multicolumn{2}{c}{\textbf{Toptee}} &\multicolumn{2}{c}{\textbf{Average}} \\

    \midrule
    
    Backbone &  Method & \textbf{R@10} & \textbf{R@50} & \textbf{R@10} & \textbf{R@50} & \textbf{R@10} & \textbf{R@50} & \textbf{R@10} & \textbf{R@50}  \\
    \midrule
    \multirow{5}{*}{ViT-G/14} & Pic2Word & 33.17 & 50.39 & 25.43 & 47.65 & 35.24 & 57.62 & 31.28 & 51.89 \\
    & SEARLE & 36.46 & 55.35 & 28.16 & 50.32 & 39.83 & 61.45 & 34.81 & 55.71 \\
    & LinCIR & 46.76 & 65.11 & 38.08 & 60.88 & \textbf{50.48} & 71.09 & 45.11 & 65.69 \\
    & \textbf{LinCIR + \ModuleName} & \textbf{48.04} & \textbf{66.68} & \textbf{39.02} & \textbf{61.03} & 50.18 & \textbf{71.14} & \textbf{45.74} & \textbf{66.28} \\
    & (vs. LinCIR) & \textcolor{darkgreen}{+1.28} & \textcolor{darkgreen}{+1.57} & \textcolor{darkgreen}{+0.94} & \textcolor{darkgreen}{+0.15} & \textcolor{red}{-0.30} & \textcolor{darkgreen}{+0.05} & \textcolor{darkgreen}{+0.63} & \textcolor{darkgreen}{+0.59} \\

    \bottomrule
  \end{tabular}
  \caption{ Quantitative results in FashionIQ dataset.}
  \label{tab:Fashion}
\end{table*}

\begin{figure*}[tb]
    \setlength{\abovecaptionskip}{-0.cm}
    \setlength{\belowcaptionskip}{-0.cm}
    \begin{center}
        \includegraphics[width=1.0\textwidth]{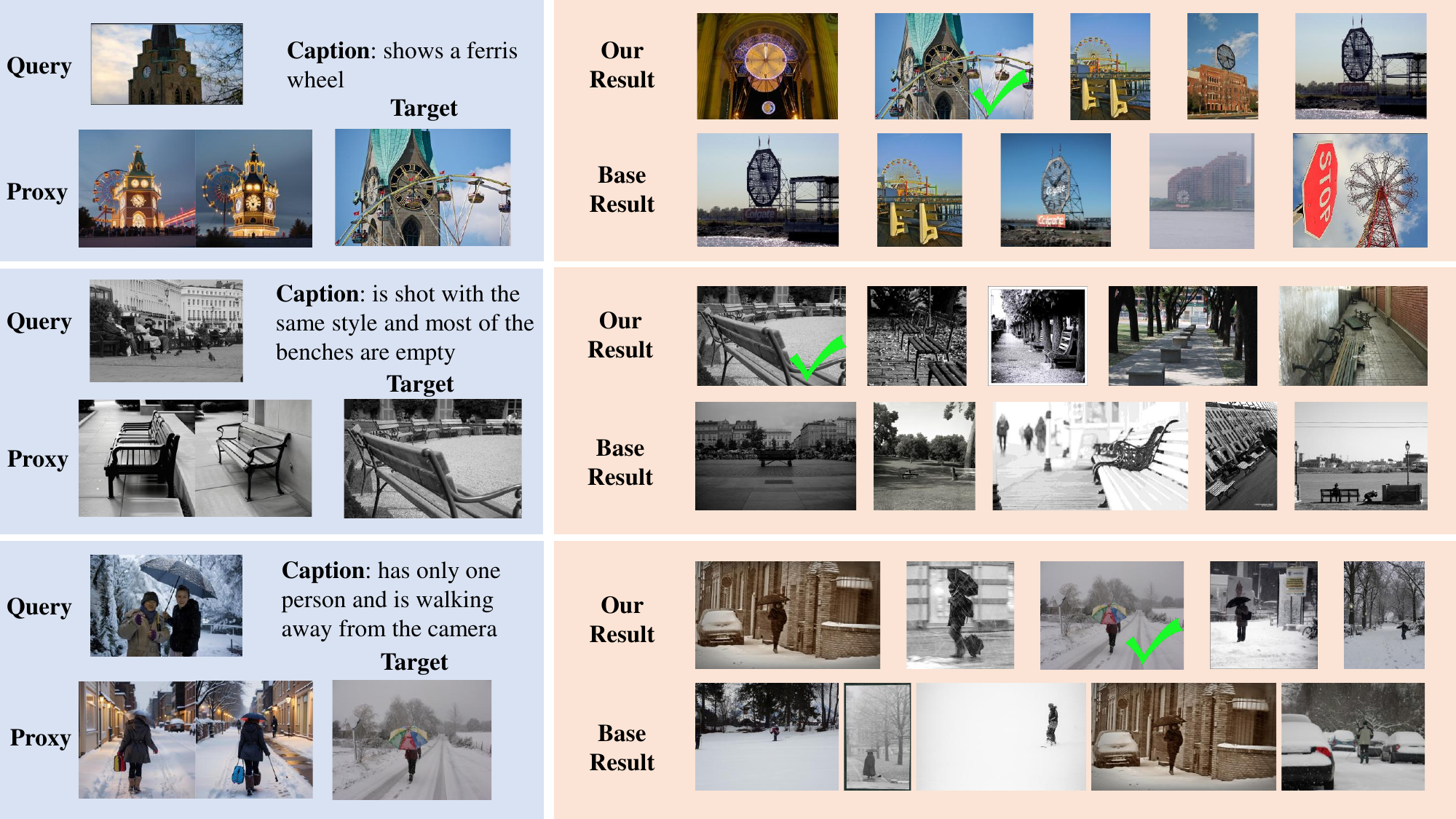}
        
    \end{center}

    \caption{ \textbf{Qualitative results of our method.} We conducted experiments on the CIRCO validation dataset to observe in which cases our method improves retrieval results. In the blue section on the left, we display the query information used for retrieval, the ground truth target image, and our generated proxy image features. `Query' represents the input query image, `Caption' represents the relative text, and in the 'Proxy' section, we show two generated imagined proxies. In the red section on the right, the top and bottom rows display the top-5 retrieval results enhanced by our imagined proxies and the baseline's top-5 retrieval results, respectively.} 
    
    \label{fig:result}
    \vspace{-5pt}
  \end{figure*}

\subsection{Balanced Proxy Retrieval}
\label{sec:ensemble}

Once the proxy image is generated, the key issue lies in how to leverage this proxy image in the retrieval. A straightforward method is to compute the proxy similarity of the retrieval database, weighted sum with the text similarity. However, as shown in the second line of Fig.\ref{fig:ablation_visual}, directly using the CLIP features of the proxy image (+PI) can lead to an excessive focus on certain elements, like the beach background. This may cause the results to overly emphasize the visual information in the image while neglecting some details from the text. So in this section, we propose to construct a robust retrieval proxy and balance the retrieval result from image and text.

\noindent \textbf{Constructing Robust Proxy.} Inspired by LDRE~\cite{ldre},  we use the LLM~\cite{gpt4,qwen} to infer a set of target captions $f_t$. With the BLIP2~\cite{blip2} generated query captions $f_o$, we are able to derive features that include semantic perturbation $f_s = f_t - f_o$, which represents a reasonable perturbation of the proxy features based on the edit direction. We believe that \textbf{introducing semantic perturbation} can help us to some extent \textbf{ignore overly emphasized information in the proxy images}. Besides, the proxy image may \textit{lose some of the query image information}, so we also incorporate the features of the query image to compensate for this loss of semantics. As shown in Fig.\ref{fig:fig_2} (b), we combine the query image’s features $f_q$, the proxy image’s feature $f_p$ with the semantic perturbation $f_s$ into the robust proxy $f_{RP}$:

\begin{equation}
f_{RP} = f_p + \frac{max(f_p)}{max(f_q)} f_q + \frac{max(f_p)}{max(f_s)} * f_s.
\end{equation}

Additionally, the weights of each component can be adjusted to control retrieval bias based on the quality of the proxy image and specific retrieval goals. For more details, refer to the supplementary materials.

\noindent \textbf{Balancing Retrieval Results.}
After obtaining the retrieval feature in text and image spaces, a critical question arises: \textit{How should we balance the two retrieval results?} The proxy features we obtain exist in the image feature space, making it difficult to directly perform weighted summation with the baseline text features. Therefore, we choose to combine their similarities. A simple approach is to take the average or weighted sum, however, in extreme cases, when we have a case with a 0.999 text similarity but only a 0.001 proxy similarity, and another case with both a 0.5 base similarity and a 0.5 proxy similarity, their averages are both 0.5, which \textit{could lead to retrieving overly extreme results}. We thus propose a balanced metric for combining these results. Specifically, given the baseline similarity $S_t$ and proxy similarity $S_p$, the final similarity$S_f$ for retrieval is as follows:

\begin{equation}
S_f = \lambda S_t + (1 - \lambda) S_b, S_b =  S_t * S_p,
\label{eq2}
\end{equation}
where $S_b$ represents the balanced similarity, $\lambda$ is a balancing parameter. We believe that the balance metric enables us to focus more on cases that perform well in both similarities.
\section{Experiments}
\label{sec:exp}
\newcommand{\imporve}{\textcolor{green}}
\newcommand{\down}{\textcolor{red}}

In this section, we demonstrate the effectiveness of our method. In Section \ref{sec:setup}, We will provide a detailed description of our experimental setup, including the baselines, evaluation metrics, and implementation details. In Section \ref{result}, we compare our method with other state-of-the-art approaches. We explore the improvements in retrieval performance by presenting some visual examples, highlighting the impact of our method in Section \ref{qualitative}, and in Section \ref{ablation}, we conduct ablation studies to investigate the contribution of each component of our method, as well as the influence of various parameter settings.

\subsection{Experimental Setup}
\label{sec:setup}

\noindent \textbf{Datasets.} We validate the performance of our method by combining them with several state-of-the-art methods on three datasets: CIRR~\cite{cirr}, CIRCO~\cite{circo}, and FashionIQ~\cite{fashioniq}. CIRR contains 21,552 real images from the NLVR~\cite{nvrl} dataset. CIRCO is built from the COCO 2017 unlabeled set~\cite{coco}, includes a validation set with 220 queries and a test set with 800 queries. FashionIQ~\cite{fashioniq} is a dataset dedicated to the fashion domain, organized into three distinct subcategories: Dress, Shirt, and Toptee. It contains 30,135 query triplets and 77,683 images available for retrieval.

\noindent \textbf{Baseline.}
We selected LDRE\cite{ldre} and LinCIR\cite{lincir} as our baselines. For each query, we generate 5 proxy images and obtain the enhanced retrieval results. We compare the improved results with some of the most recent ZS-CIR results, such as Searle\cite{searle}, CIReVL\cite{vbl}.

\noindent \textbf{Evaluation Metrics.} 
In \textbf{CIRCO}, each query has multiple target images, so we calculate mAP@k to offer a more detailed assessment, where $k \in \{5, 10, 25, 50\} $ indicates the number of top-ranked retrieval results.
For \textbf{CIRR}, we follow the original benchmarks and use Recall@k (with $k \in \{1, 5, 10, 50\}$ ) as the primary metric. We also assess performance in a subset setting, which we denote as Recall$_{Subset}$@K (with $k \in \{1, 2, 3\}$  ). On \textbf{FashionIQ}, we also use Recall@K (with $k \in \{10, 50\}$) for evaluation.

\noindent \textbf{Implement Details.}
We primarily conducted experiments using two backbone models: CLIP-L and CLIP-G. We used Qwen1.5-32B\cite{qwen} as the LLM in all the experiments, which yielded results comparable to those reported in the original papers.  For generating proxy images, we use MIGC\cite{migc} with SD1.5 as the backbone. We use Qwen1.5-32B as our LLM to generate the layout for each query and then generate 5 proxy images based on the query image and layout. All the experiments are conducted by Pytorch with one NVIDIA A6000 GPU.

\begin{table*}[!htb]
\setlength\tabcolsep{6pt}
\setlength{\abovecaptionskip}{0.cm}
    \setlength{\belowcaptionskip}{0.cm}
  \centering

  \begin{tabular}{c c c c | c c c c c | c c c c}
    \toprule
    \multicolumn{4}{c|}{\textbf{Benchmark}} & \multicolumn{5}{c|}{\textbf{CIRR}} & \multicolumn{4}{c}{\textbf{CIRCO}} \\

    \midrule
    
    \multicolumn{4}{c|}{\textbf{Metric}}  & \multicolumn{5}{c|}{\textbf{Recall@K}} & \multicolumn{4}{c}{\textbf{mAP@K}}  \\
    \midrule
    
     Backbone &  PI & RR & BM  & k=1 & k=2 & k=5 &  k=10 & k=50 & k=5 & k=10 & k=25 & k=50 \\
    \midrule
    \multirow{7}{*}{LDRE-G} & & & & 36.15 & 49.49 & 66.39 & 77.25 & 93.95 & 31.12 & 32.24 & 34.95 & 36.03 \\
     & \multirow{2}{*}{\checkmark} & & & \cellcolor{gray!30}36.02  & \cellcolor{gray!30}50.02 & \cellcolor{gray!30}66.51  & \cellcolor{gray!30}77.52  & \cellcolor{gray!30}93.74  & \cellcolor{gray!30}31.06  & \cellcolor{gray!30}32.41  & \cellcolor{gray!30}35.19  & \cellcolor{gray!30}36.35  \\
     
     & & & & \cellcolor{gray!30}\textcolor{red}{-0.13} & \cellcolor{gray!30}\textcolor{darkgreen}{+0.53} & \cellcolor{gray!30}\textcolor{darkgreen}{+0.12} & \cellcolor{gray!30}\textcolor{darkgreen}{+0.27} & \cellcolor{gray!30}\textcolor{red}{-0.21} & \cellcolor{gray!30}\textcolor{red}{-0.06} & \cellcolor{gray!30}\textcolor{darkgreen}{+0.17} & \cellcolor{gray!30}\textcolor{darkgreen}{+0.24} & \cellcolor{gray!30}\textcolor{darkgreen}{+0.32} \\
     
     & \multirow{2}{*}{\checkmark} & \multirow{2}{*}{\checkmark} & & \cellcolor{gray!10}38.15  & \cellcolor{gray!10}52.53 & \cellcolor{gray!10}68.55  & \cellcolor{gray!10}79.47  & \cellcolor{gray!10}94.48  & \cellcolor{gray!10}31.92  & \cellcolor{gray!10}33.59  & \cellcolor{gray!10}36.26  & \cellcolor{gray!10}37.37   \\

    & & & & \cellcolor{gray!10}\textcolor{darkgreen}{+2.00} & 
    \cellcolor{gray!10}\textcolor{darkgreen}{+3.04} & 
    \cellcolor{gray!10}\textcolor{darkgreen}{+2.16} & \cellcolor{gray!10}\textcolor{darkgreen}{+2.22} & \cellcolor{gray!10}\textcolor{darkgreen}{+0.53} & \cellcolor{gray!10}\textcolor{darkgreen}{+0.80} & \cellcolor{gray!10}\textcolor{darkgreen}{+1.35} & \cellcolor{gray!10}\textcolor{darkgreen}{+1.31} & \cellcolor{gray!10}\textcolor{darkgreen}{+1.34} \\
     
     & \multirow{2}{*}{\checkmark} & \multirow{2}{*}{\checkmark} & \multirow{2}{*}{\checkmark} & \cellcolor{gray!30}\textbf{39.25}  & \cellcolor{gray!30}\textbf{52.94} & \cellcolor{gray!30}\textbf{70.07} & \cellcolor{gray!30}\textbf{80.00} & \cellcolor{gray!30}\textbf{94.89} & \cellcolor{gray!30}\textbf{32.75} & \cellcolor{gray!30}\textbf{34.26} & \cellcolor{gray!30}\textbf{36.86} & \cellcolor{gray!30}\textbf{38.03} \\

     & & & & \cellcolor{gray!30}\textcolor{darkgreen}{+3.10} & \cellcolor{gray!30}\textcolor{darkgreen}{+3.45} & \cellcolor{gray!30}\textcolor{darkgreen}{+3.68} & \cellcolor{gray!30}\textcolor{darkgreen}{+2.75} & \cellcolor{gray!30}\textcolor{darkgreen}{+0.94} & \cellcolor{gray!30}\textcolor{darkgreen}{+1.63} & \cellcolor{gray!30}\textcolor{darkgreen}{+2.02} & \cellcolor{gray!30}\textcolor{darkgreen}{+1.91} & \cellcolor{gray!30}\textcolor{darkgreen}{+2.00} \\

    \bottomrule
  \end{tabular}
  \caption{ Ablation results in CIRR and CIRCO dataset.}
  \label{tab:ablation_cirr}
\end{table*}

\begin{figure}[tb]
    \setlength{\abovecaptionskip}{-0.cm}
    \setlength{\belowcaptionskip}{-0.cm}
    \begin{center}
        \includegraphics[width=0.48\textwidth]{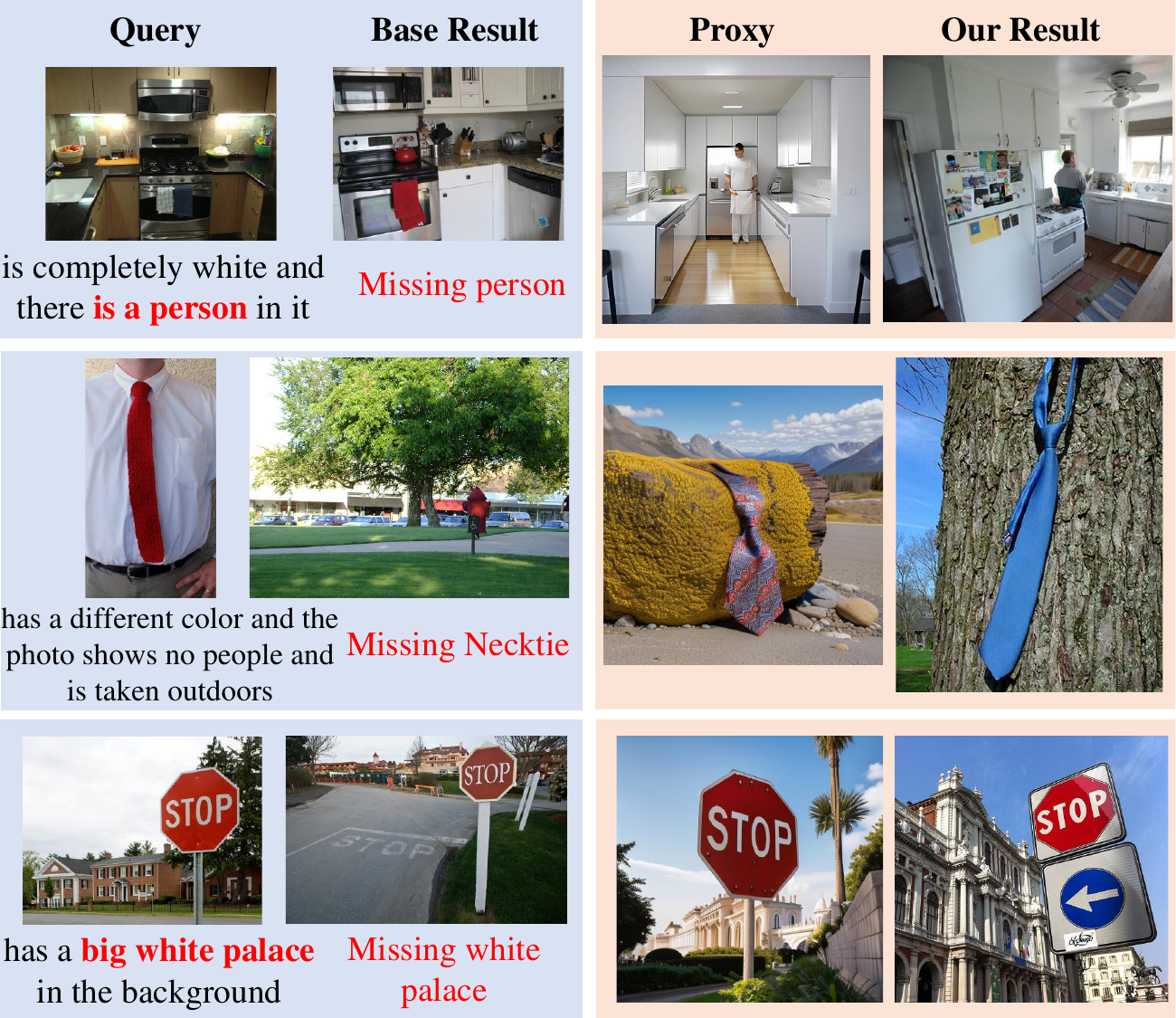}
        
    \end{center}

    \caption{ \textbf{Qualitative results of our method.} We present the result on CIRCO validation dataset with top-1 retrieval results. Baseline methods in text-based retrieval overlook certain attributes and details in complex related texts, resulting in suboptimal retrieval outcomes. In contrast, our approach can construct suitable proxies on the image side, supplementing this missing information and yielding improved retrieval results.} 
    
    \label{fig:visual_result}
    \vspace{-5pt}
  \end{figure}

\subsection{Quantitative Results}
\label{result}

\noindent \textbf{CIRCO.} We present the experimental results on the CIRCO test dataset on LDRE\cite{ldre} and LinCIR\cite{lincir}. In the left half of Tab.\ref{tab:ZSCIR}, we observe that our proxy image can improve the retrieval performance, which an approximately 5-point improvement in mAP at k=5 in LinCIR, and about 2-point improvement in LDRE. These results effectively demonstrate the validity and impact of our method.

\noindent \textbf{CIRR.} In the right half of Tab.\ref{tab:ZSCIR}, we present the experimental results on the CIRR test dataset. Our proxy image significantly improved the LDRE and LinCIR performance. The performance improvements across different methods and backbones indicate that our approach can be applied to various text-based retrieval methods.

\noindent \textbf{FashionIQ.} In Tab.\ref{tab:Fashion}, we present the results on the FashionIQ val dataset. Since LDRE has not open-sourced the code for the FashionIQ, we used LinCIR as the baseline. Although we observe a drop in R@10 in the Toptee, our method improves the retrieval accuracy in the Shirt and dress Subset, thus improving the average performance.

\begin{figure}[tb]
    \setlength{\abovecaptionskip}{-0.cm}
    \setlength{\belowcaptionskip}{-0.cm}
    \begin{center}
        \includegraphics[width=0.42\textwidth]{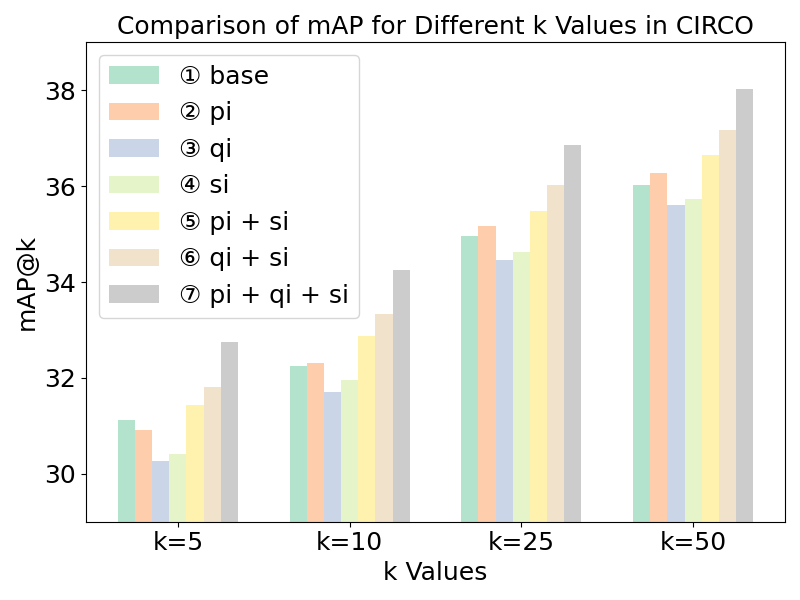}
        
    \end{center}

    \caption{ \textbf{Ablation results on the composition of robust proxy features} in the CIRCO dataset. $pi$ indicates proxy features, $si$ represents semantic perturbation, and $qi$ indicates the query features.} 
    
    \label{fig:ablation_robust}
    \vspace{-5pt}
  \end{figure}

\subsection{Qualitative Results}
\label{qualitative}

We conducted experiments on the CIRCO validation dataset to verify the improvements our method brings to retrieval.

\begin{figure*}[tb]
    \setlength{\abovecaptionskip}{-0.cm}
    \setlength{\belowcaptionskip}{-0.cm}
    \begin{center}
        \includegraphics[width=1.0\textwidth]{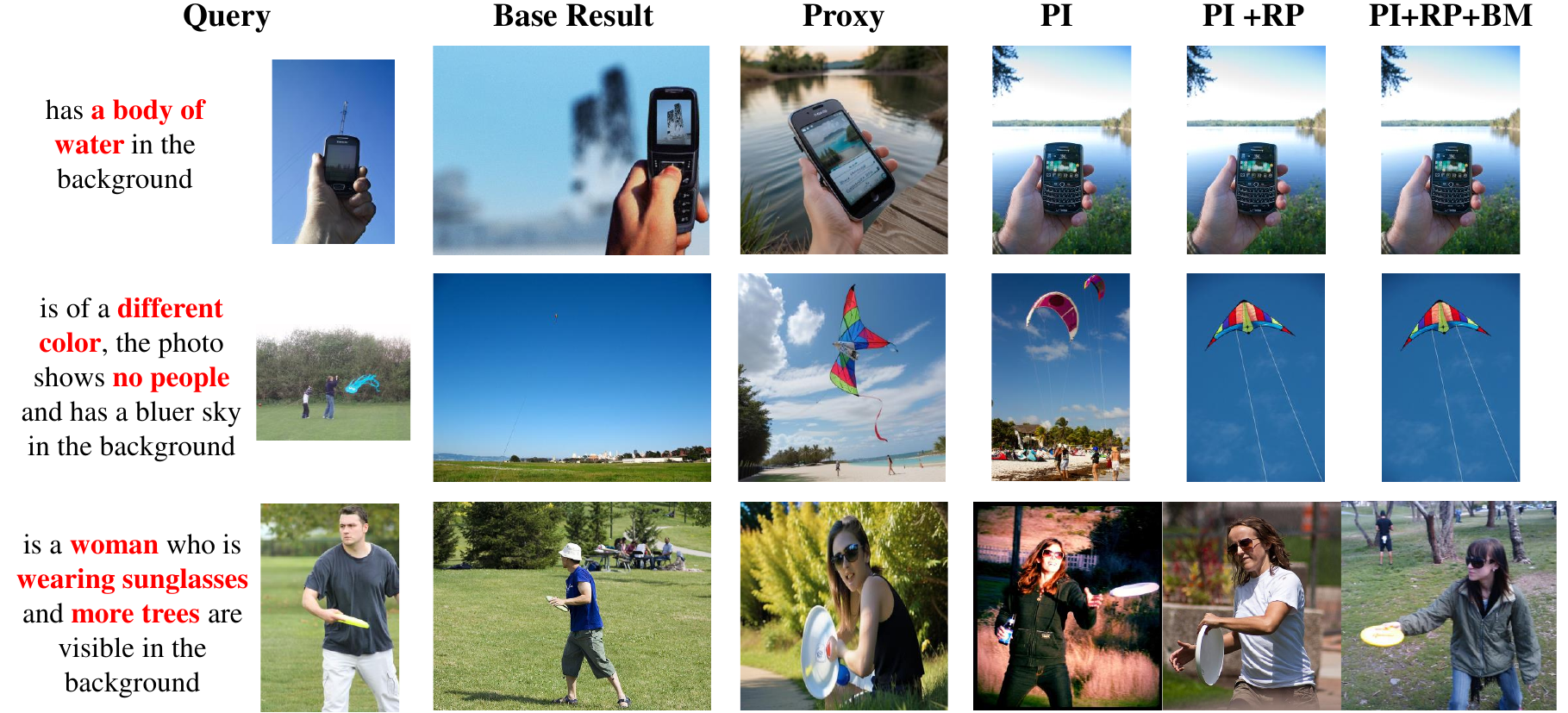}
        
    \end{center}

    \caption{ \textbf{Visualize of Ablation Result.} We present the ablation result on CIRCO dataset.} 
    
    \label{fig:ablation_visual}
  \end{figure*}

\noindent \textbf{Experiment setting.} Specifically, we used LDRE as the baseline and generated 5 imagined proxy images for each query. We present two types of results. First, based on the ground truth for each query, we show in Fig.\ref{fig:result} that our method improves the accuracy of ground truth occurrence in top-5 retrieval results. Second, disregarding the ground truth image, we demonstrate in Fig.\ref{fig:visual_result} that our method retrieves a more suitable image in the top-1 result.

\noindent \textbf{Analysis of Experimental Results on TOP-5 Retrieval.} From the visualized results in Fig.\ref{fig:result}, we can draw the following conclusions: (1) Our generated proxy images effectively preserve the style and key elements of the original image. In the first row, our image maintains the clock tower as a significant element and adheres to the related caption by generating a Ferris wheel. In the second row, our image retains the original black-and-white style. (2) The imagined proxies complement important information that may be overlooked by text alone. For instance, in the first-row example, the clock tower—a crucial concept—is missing in the text-only retrieval result, while our imagined proxy and the original image both provide this retrieval information. In the third row, our constructed image improves retrieval accuracy by incorporating an umbrella and snow.

\noindent \textbf{Analysis of Experimental Results on TOP-1 Retrieval.} Cases in Fig.\ref{fig:visual_result} show that by providing proxy image information, our approach achieves more relevant retrievals. \textbf{Retrieval based solely on text space may only capture partial information in complex contexts}, leading to the \textbf{omission of certain details} in the images—such as the person in the first row, the tie in the second row, and the White House in the background of the third row. Our method incorporates these details in the images, supplementing and balancing this information to yield more suitable results.

\begin{figure*}[tb]
    \setlength{\abovecaptionskip}{-0.cm}
    \setlength{\belowcaptionskip}{-0.cm}
    \begin{center}
        \includegraphics[width=01.0\textwidth]{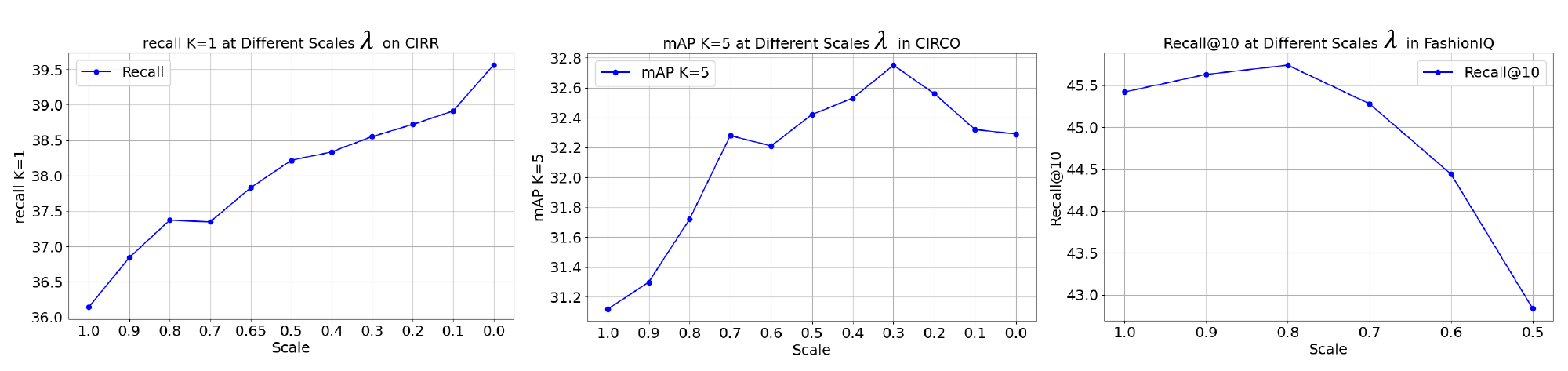}
        
    \end{center}

    \caption{ \textbf{The impact of weight $\lambda$} in CIRCO, CIRR, and FashionIQ. As $\lambda$ decreases, the proportion of proxy information increases. } 
    
    \label{fig:ablation_hp}
    \vspace{-5pt}
  \end{figure*}

\subsection{Ablation Study}
\label{ablation}

\noindent \textbf{Effects of the Proxy Image Features.} We first conduct an ablation study on the proposed proxy image features in the CIRR dataset and CIRCO dataset. 
We use PI (Proxy Image) refers to the approach that directly computes retrieval similarity from proxy image features, and then performs a weighted sum with the original similarity. RP (Robust Proxy) indicates integrating the proxy image into more robust proxy features. BM (Balance Metric) refers to using the balance metric. The results from the CIRR and CIRCO are shown in Tab.\ref{tab:ablation_cirr}. We can draw the following conclusions: 1) The imagined proxy images may contain some noise, leading to the retrieval focus on irrelevant details. we observe a decrease in recall at k=1 and 50 in CIRR, and mAP k=5 decreases in CIRCO. 2) After integrating the proxy image into a more robust proxy feature, we see a significant improvement in both datasets, which demonstrates the effectiveness of our robust proxy features. 3) By introducing the balance metric between proxy and text similarity, we further improve the retrieval performance. This suggests that combining the information from both image and text domains leads to more accurate retrieval results. 

We can draw similar conclusions from Fig.\ref{fig:ablation_visual}, which shows retrieval results on the CIRCO test dataset. In the first row, the proxy image provides a realistic and accurate depiction of the target image, allowing for improved retrieval results. In the second row, relying solely on the query image may cause the retrieval to overly focus on the beach background, contradicting the ``no people" text information. Here, using the robust proxy (PI+RP) further enhances the overlooked aspects. In the third row, the balance metric balances the `the woman with sunglasses' in proxy features with `trees' in baseline features, achieving better retrieval results for both the subject and the background.

\begin{figure}[tb]
    \setlength{\abovecaptionskip}{-0.cm}
    \setlength{\belowcaptionskip}{-0.cm}
    \begin{center}
        \includegraphics[width=0.35\textwidth]{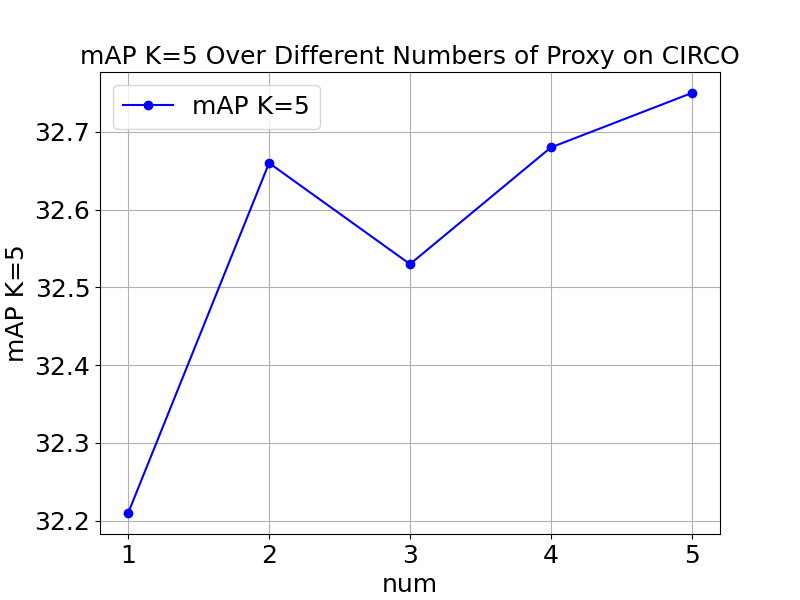}
        
    \end{center}

    \caption{ \textbf{Parameter Analysis.} The effect of varying numbers of proxy images on performance for the CIRCO dataset.} 
    
    \label{fig:ablation_num}
    \vspace{-5pt}
  \end{figure}

\noindent \textbf{Ablation on Robust Features.} We analyze the composition of our robust proxy on the CIRCO dataset, with results shown in Fig.\ref{fig:ablation_robust}. The robust proxy is composed of three main components: $q_i$ represents the query image features, $s_i$ represents the semantic perturbation, and $p_i$ represents our proxy image. A comparison among \ding{172}, \ding{173}, \ding{174}, and \ding{175} shows that directly using our proxy image features can enhance retrieval performance to a certain extent, whereas relying solely on the query image or semantic perturbation tends to focus excessively on incorrect information, leading to a decline in performance. Additionally, comparisons among \ding{176}, \ding{177}, and \ding{178} demonstrate that combining all three features together improves retrieval accuracy.

\noindent \textbf{Effects of the weight $\lambda$.} We analyze the $\lambda$ used in Eq.~(\ref{eq2}). The experiments revealed that differences in dataset properties, result in different optimal values for $\lambda$. As shown in Fig.\ref{fig:ablation_hp}, we found that the suitable $\lambda$ for CIRCO, CIRR, and FashionIQ are around 0.3, 0.0, and 0.8.

\noindent \textbf{Effects of the number of proxy images.} We tested the impact of using different numbers of proxy images on performance with the CIRCO dataset. As shown in Fig.\ref{fig:ablation_num}, retrieval performance improves as the number of proxy images increases, though the rate of improvement gradually slows down. Therefore, the number of proxy images can be chosen based on a balance between efficiency and accuracy.

\section{Conclusion}
\label{sec:conclusion}

In this paper, we propose the \ModuleName, a plug-and-play training-free method for current ZSCIR  tasks. Our method imagines proxy images, which align with the query image and the relative captions, providing fine-grained details that may be overlooked by the text. We combine the proxy images, query images, and semantic perturbation into more robust proxy image features, and propose a balance metric to compose the proxy and baseline similarities, enhancing the retrieval performance. The experiment results in CIRR, CIRCO, and FashionIQ datasets show that our method can successfully improve retrieval accuracy and demonstrate the potential for providing additional retrieval information from the image side, revealing a new direction for improving retrieval accuracy in the future.

{
    \small
    \bibliographystyle{ieeenat_fullname}
    \bibliography{main}
}

\clearpage

\setcounter{page}{1}
\setcounter{section}{0}
\setcounter{figure}{0}
\setcounter{section}{0}
\setcounter{equation}{0}
\gdef\thesection{\Alph{section}}
\maketitlesupplementary

\begin{figure}[!tb]
    \setlength{\abovecaptionskip}{-0.cm}
    \setlength{\belowcaptionskip}{-0.cm}
    \begin{center}
        \includegraphics[width=0.48\textwidth]{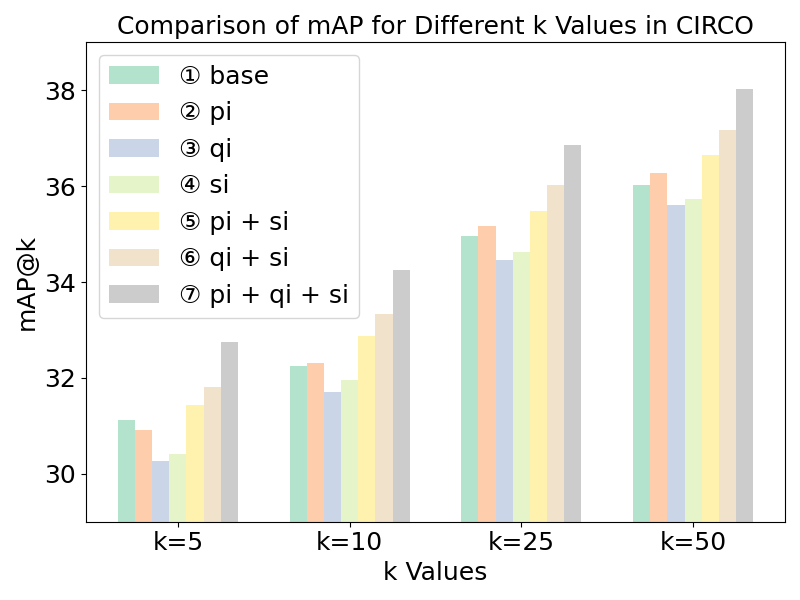}
        
    \end{center}

    \caption{ \textbf{Ablation results on the composition of robust proxy features} in the CIRCO dataset. \textbf{pi} indicates proxy features, \textbf{si} represents semantic perturbation, and \textbf{qi} indicates the query features. } 
    
    \label{fig:rp}
    \vspace{-5pt}
  \end{figure}

\begin{figure*}[!tb]
    \setlength{\abovecaptionskip}{-0.cm}
    \setlength{\belowcaptionskip}{-0.cm}
    \begin{center}
        \includegraphics[width=1.0\textwidth]{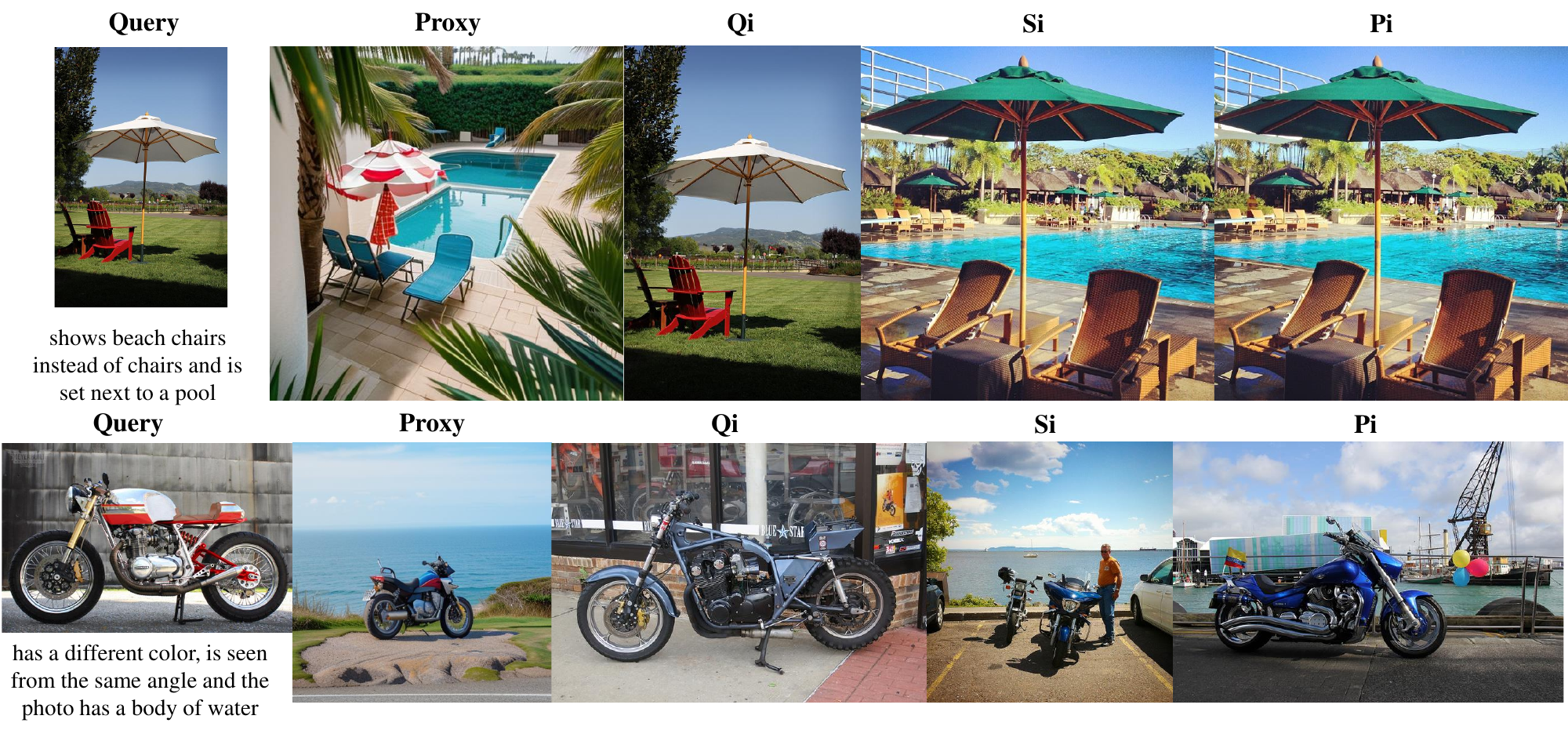}
        
    \end{center}

    \caption{ \textbf{Ablation result on the Robust Proxy.} We present the visualization result of using different compositions (\textbf{Qi} represents only using query image, \textbf{Si} represents only semantic perturbation, and \textbf{Pi} represents only using the proxy image) of features in the robust proxy.}
    
    \label{fig:supp_rp2}
    \vspace{-5pt}
  \end{figure*}

\section{More Analysis}

In this section, we will conduct a deeper analysis of the experiments to better illustrate the details of our method.

\subsection{The Construction of Robust Proxy}

We analyze \textbf{the specific roles of each component in the robust features through Fig.\ref{fig:rp} and more visual examples in Fig.\ref{fig:supp_rp2}}. The RP features include proxy image features, query image features, and semantic perturbation. Using the CIRCO dataset as an example, we construct RP features with different components for retrieval and obtain the top-1 retrieved image. This allows us to examine how varying the RP features influences the retrieved images.

We can draw the following conclusions: \textbf{1) The proxy image can provide richer information.}  Since the proxy image contains semantic editing information as well as some information from the query image, using the proxy image can offer more comprehensive information. In Fig.\ref{fig:rp}, compared to the baseline retrieval results \ding{172},  \ding{173} improves retrieval precision at $k=10,25,50$, although it slightly reduces precision at $k=5$. In contrast, directly using semantic perturbation or query features leads to an overall decline in retrieval precision. It can also be observed in Fig.\ref{fig:supp_rp2} that using only the query image (Qi) results in \textbf{an incorrect emphasis on background, even directly retrieving the original image itself.} While the second row shows that using only semantic perturbation (Si) \textbf{ignores details in the image such as angle}. In contrast, the proxy image achieves relatively better retrieval results because it provides background and angle information. \textbf{2) All the features are important.} In Fig.\ref{fig:rp}, compared to \ding{173} and \ding{174}, when semantic perturbation is applied to enhance the text-driven editing information in proxy or query image features (as shown in \ding{176} and \ding{177}), retrieval precision improves, highlighting the importance of semantic perturbation for robust features. Besides, the comparisons between \ding{176} and \ding{178}, as well as \ding{177} and \ding{178}, demonstrate that adding either proxy image features or query image features can further enhance the effectiveness of robust features. The second column in Fig.\ref{fig:supp_fashioniq} indicates that combining query image features can better preserve certain textual patterns and textures that are challenging for the proxy image to generate accurately.

\begin{figure*}[!tb]
    \setlength{\abovecaptionskip}{-0.cm}
    \setlength{\belowcaptionskip}{-0.cm}
    \begin{center}
        \includegraphics[width=1.0\textwidth]{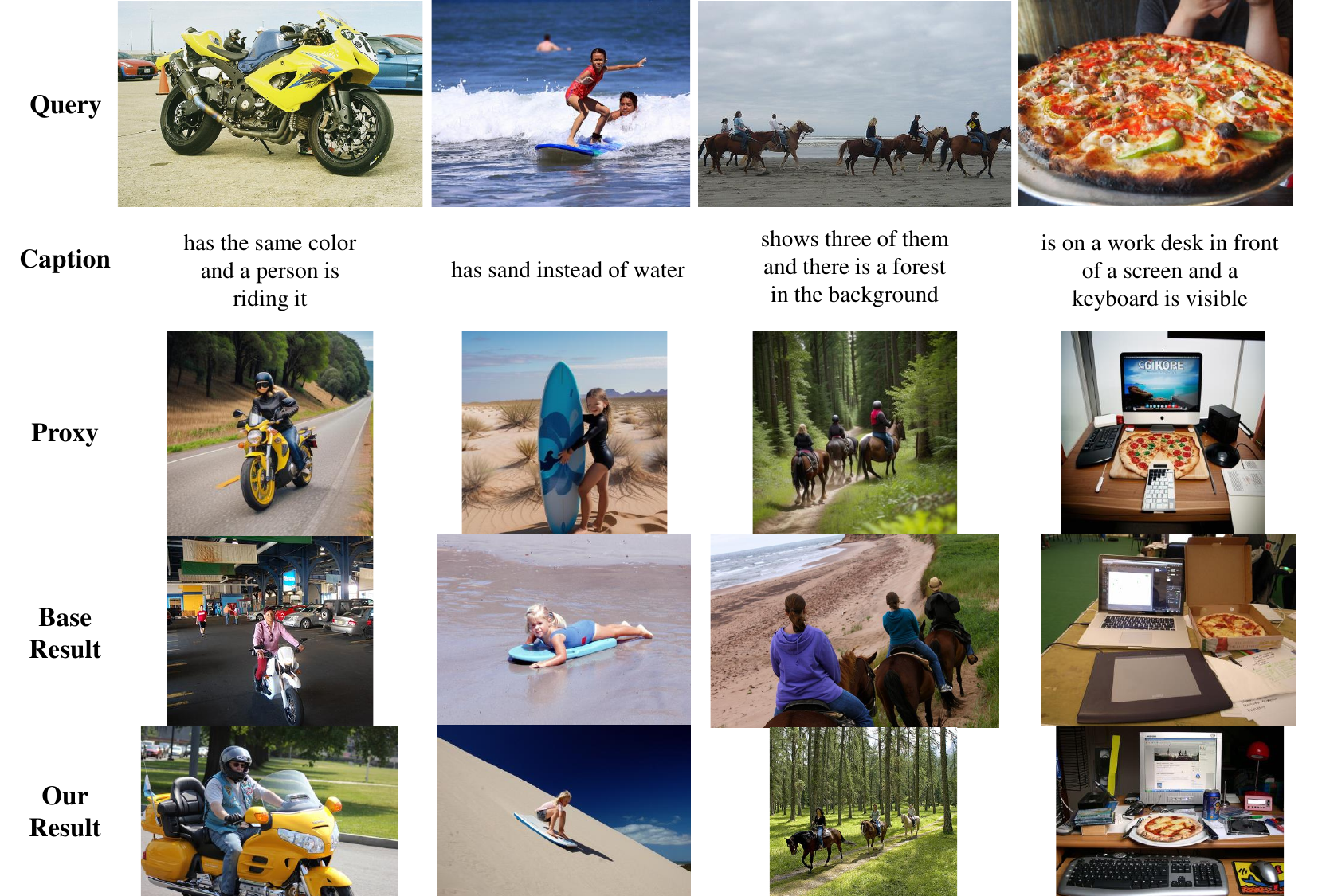}
        
    \end{center}

    \caption{ \textbf{Qualitative results on CIRCO dataset.} We show more improvement in top-1 retrieval results in the CIRCO dataset. } 
    
    \label{fig:supp_circo}
    \vspace{-5pt}
  \end{figure*}

\begin{figure*}[tb]
    \setlength{\abovecaptionskip}{-0.cm}
    \setlength{\belowcaptionskip}{-0.cm}
    \begin{center}
        \includegraphics[width=0.9\textwidth]{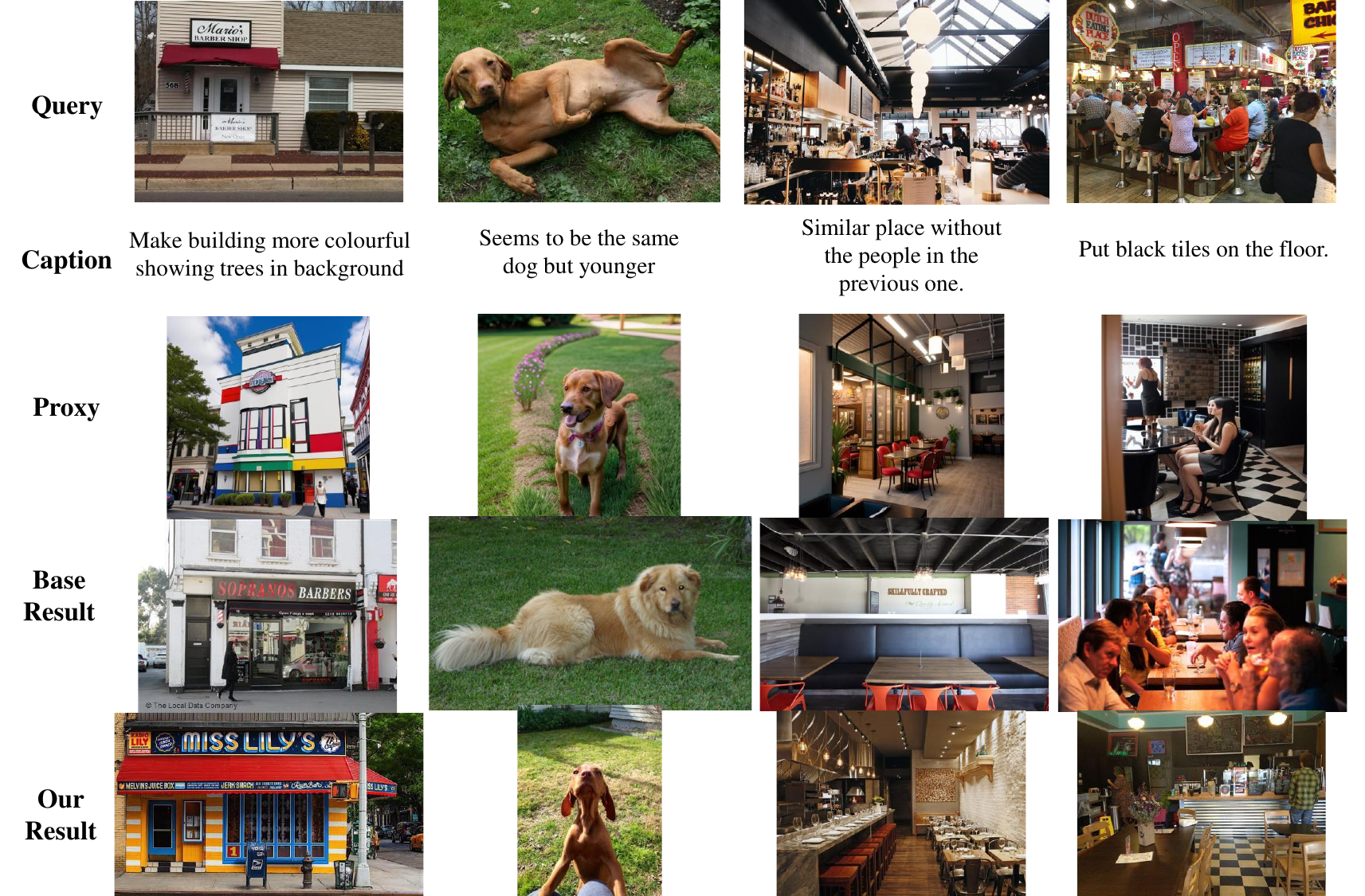}
        
    \end{center}

    \caption{ \textbf{Qualitative results on CIRR dataset.} We show more improvement in top-1 retrieval results in the CIRR dataset.} 
    
    \label{fig:supp_cirr}
    \vspace{-5pt}
  \end{figure*}

\subsection{The analysis of hyperparameters.}

Since object content, associated text formats, and the alignment between text and target images vary across datasets, different datasets require different weighting parameters $\lambda$ during retrieval. Additionally, the quality of proxies constructed for different datasets may also vary. For example, we observed that in FashionIQ, \textbf{the constructed proxies struggle to fully describe corresponding text for logo patterns and face challenges in generating pure white backgrounds during the generation process}. As a result, the role of proxy images in FashionIQ is relatively weaker, necessitating a larger $\lambda$.
Moreover, if different backbones are used, the extracted features may emphasize different attributes, requiring dataset-specific adjustments to the $\lambda$ parameter during retrieval.

Users can refine robust features by adjusting feature weights to emphasize specific retrieval aspects. For example, increasing the weight of the query image enhances details like logos or highlights attributes such as angle or style. Conversely, emphasizing proxy images or semantic perturbations shifts focus toward text-based editing directions.

\section{More Qualitative Results}

In this section, we show more results on three datasets. We present the Query image, relative caption, one of our generated Proxy images, as well as the top-1 baseline retrieval result and top-1 retrieval result of our method.

\subsection{More Qualitative results on CIRCO.} We show qualitative results on CIRCO in Fig.\ref{fig:supp_circo}. We can draw the following conclusions: 1) The generated proxy images demonstrate certain detailed features, such as the yellow car in the first column, the desert and blue skateboard in the second column, the forest background and three horses in the third column, and the screen and keyboard in the last column. 2) Our method improves the Top-1 retrieval results to better meet the requirements to some extent. For example, in the first column, the car we retrieved is yellow. In the second column, the background of our result has sand instead of water. In the third column, with a proxy to imagine the forest scene, we successfully retrieved three horses in a forest. In the last column, the retrieved result closely matches the features presented by the proxy and better fits the described spatial characteristics.

\subsection{More Qualitative results on CIRR.} We show more qualitative results on CIRR in Fig.\ref{fig:supp_cirr} with the improvements in TOP-1 retrieval performance. As shown, our generated proxy images provide elements such as the target's background ambiance and scene (e.g., more colorful background and black tiles), as well as the type and fine-grained attributes of the main objects (e.g., the same type of dog). These enhancements contribute to improved retrieval accuracy.

\begin{figure*}[tb]
    \setlength{\abovecaptionskip}{-0.cm}
    \setlength{\belowcaptionskip}{-0.cm}
    \begin{center}
        \includegraphics[width=0.9\textwidth]{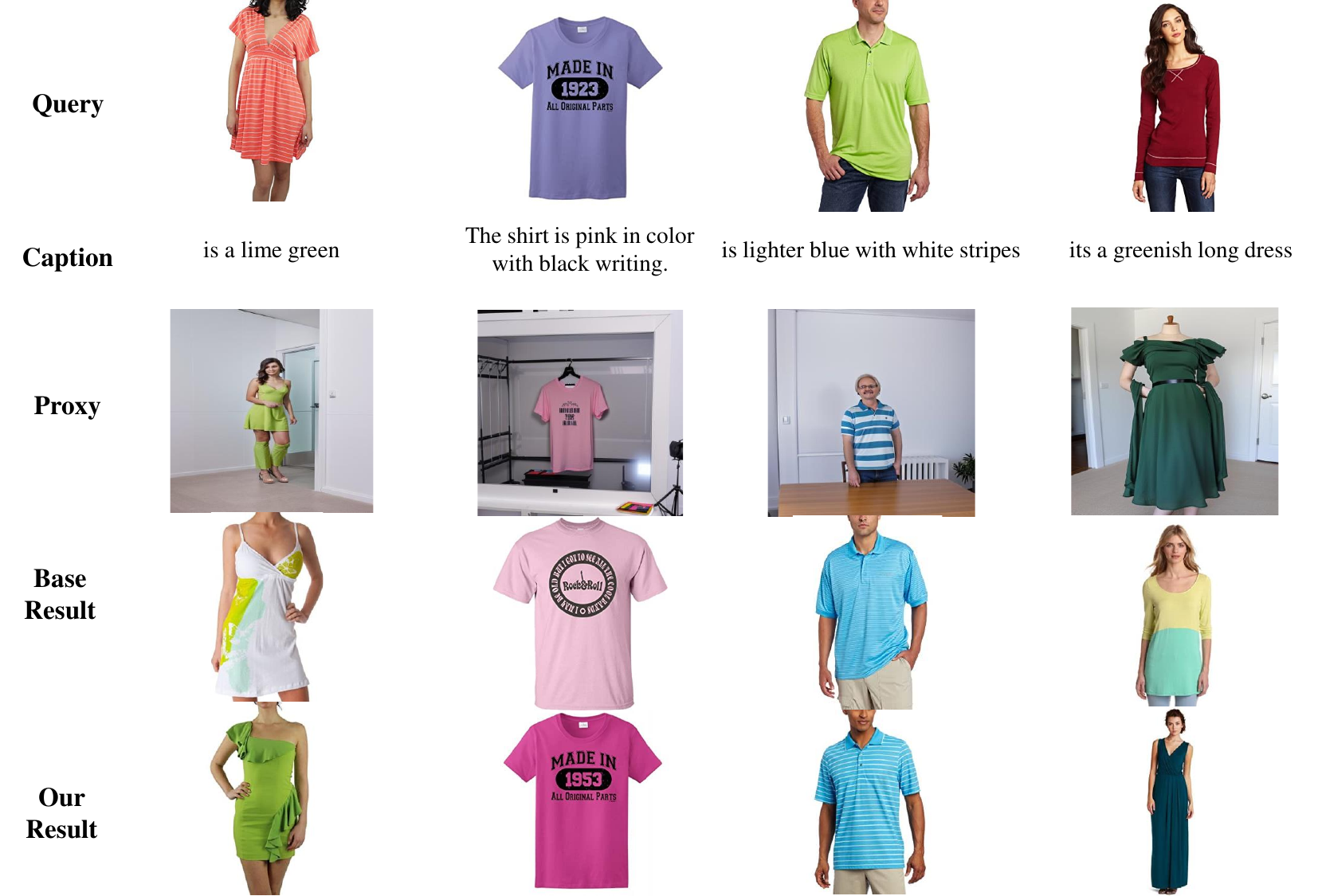}
        
    \end{center}

    \caption{ \textbf{Qualitative results on FashionIQ dataset.} We show more improvement in top-1 retrieval results in the FashionIQ dataset.} 
    
    \label{fig:supp_fashioniq}
    \vspace{-5pt}
  \end{figure*}

\subsection{More Qualitative results on FashionIQ.} We show more qualitative results on FashionIQ in Fig.\ref{fig:supp_fashioniq} with the improvements in TOP-1 retrieval performance. 
FashionIQ provides text descriptions of attributes such as color and patterns, enabling our proxy features to achieve relatively better retrieval results. At the same time, we believe that query image features are also important for the FashionIQ dataset. For example, in the second column, although the baseline can retrieve clothes with pink color and black text, incorporating original image features allows the retrieval of images with logos that are more similar to the query while also aligning with the text description.

\section{Limitation}

\noindent \textbf{Additional time overhead.} While our method is plug-and-play in most scenarios and improves retrieval accuracy, it does introduce some time overhead. The process of layout generation and image generation adds certain time costs. 

\noindent \textbf{Sensitive to hyperparameters.} The image-based retrieval enhancement is influenced by the performance of the constructed proxy images and the trade-off parameters used. Users need to carefully set reasonable weighting hyperparameters for the retrieval process. Thus, 
how to design a more reasonable method for combining weights or metrics to reduce sensitivity to parameters is an important direction for future research.

\end{document}